# Re-presenting a Story by Emotional Factors using Sentimental Analysis Method

**Hwiyeol Jo (hwiyeolj@gmail.com)**
Department of Computer Science & Engineering, Seoul National University,
Seoul, 151-742, Republic of Korea

**Yohan Moon (ttattang11@naver.com)**
Department of Interactional Science, Sungkyunkwan University
Seoul, 110-745, Republic of Korea

**Jong In Kim (prows12@gmail.com)**
Interdisciplinary Program in Cognitive Science, Seoul National University
Seoul, 151-742, Republic of Korea

**Jeong Ryu (ryujeong@gmail.com)**
Department of Psychology, Yonsei University
Seoul, 120-749, Republic of Korea

**Abstract**

Remembering an event is affected by personal emotional status. We examined the psychological status and personal factors; depression (Center for Epidemiological Studies – Depression, Radloff, 1977), present affective (Positive Affective and Negative Affective Schedule, Watson et al., 1988), life orient (Life Orient Test, Scheier & Carver, 1985), self-awareness (Core Self Evaluation Scale, Judge et al., 2003), and social factor (Social Support, Sarason et al., 1983) of undergraduate students (N=64) and got summaries of a story, Chronicle of a Death Foretold (Gabriel García Márquez, 1981) from them. We implement a sentimental analysis model based on convolutional neural network (LeCun & Bengio, 1995) to evaluate each summary. From the same vein used for transfer learning (Pan & Yang, 2010), we collected 38,265 movie review data to train the model and then use them to score summaries of each student. The results of CES-D and PANAS show the relationship between emotion and memory retrieval as follows: depressed people have shown a tendency of representing a story more negatively, and they seemed less expressive. People with full of emotion – high in PANAS - have retrieved their memory more expressively than others, using more negative words then others.

The contributions of this study can be summarized as follows: First, lightening the relationship between emotion and its effect during times of storing or retrieving a memory. Second, suggesting objective methods to evaluate the intensity of emotion in natural language format, using a sentimental analysis model.

## Introduction

"How do people memorize an event" has been a long lasting question suggested in various fields of study including Philosophy, Psychology, Cognitive Science, Brain Science and most importantly, Artificial Intelligence. In the case of Artificial Intelligence, some researchers have been interested in methodology, particularly in efficient storage of an event into a computer and the retrieval of the event. Others who have focused on implementing human-like intelligence have wanted to imitate such retrieval process of human. Such system has to be able to retrieve the memorized facts and add their feelings to the memories simultaneously. The first group of researchers should find efficient representation method to embed each and every events occurring on our surroundings into computer, and the second group should develop an advanced type of representation capable of other various information.

In this study, being on the second point of view, we investigate the relationship between emotion and the polarity and intensity of the retrieved memory. As an experiment design, we introduce a novel to give participants the same experience, which and then evaluate their summaries using sentimental analysis model trained by 38,265 movie review data. Then we've analyzed the results with their emotional status. The overall architecture of our system is as follows:

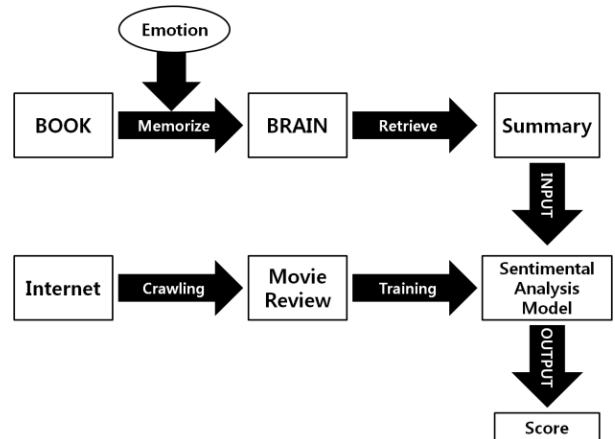

Figure 1: Overall framework

As a secondary goal, we've suggested a new methodology to evaluate emotion. Research which studied the emotion try to assess the emotion using only measurements (e.g.

PANAS), otherwise determine is by subjective methods. Mayer & Geher (1996) use group consensus to identify the emotion, Diener et al. (1985) computes the intensity of emotion by the difference between positive scale in a day and mean positive scale. Many of recent studies were unsuccessful in representing a method to determine what the emotion and how its intensity objectively without measurements likes PANAS.

The contributions of this study can be summarized as follows: First, lightening the relationship between emotion and its effect during times of storing or retrieving a memory. Second, suggesting objective methods to evaluate the intensity of emotion in natural language format, using a sentimental analysis model.

To show the difference in retrieving a memory under equal circumstance, we've used the 'Chronicle of a Death Foretold.' The novel is used in various field of study, ranging from literature and language (Aghaei & Hayati, 2014), cognitive science (Popova, 2015), psychology (Scheper-Hughes, 2003), culture (Weiss, 1990) and so forth.

We will briefly summarize the background knowledge needed for understanding our method, which consists of our experimental data and training data of sentimental analysis model each other. Afterwards, we will introduce our overall framework to represent the relationship, explain the results, and draw a conclusion and discussion with derived results.

## Backgrounds

### Emotion effects on Memory

Emotion controls the contents of memories both when we store and retrieve them. A theory "Affect as information framework" explains that if we are positive, we tend to be interpretive, relational processing while being detailed, stimulus bound, referential processing in negative affective (Clore et al., 2001). Kensinger et al. (2007) found the different specificity of an event.

Emotion can consolidate memory. Emotional memories are kept fairly untouched and forgotten slowly because they were rehearsed again and again. The negative emotion can improve an accuracy of the memory (Kensinger, 2007). A research showed that emotion can make affect regulation (Raes et al., 2003).

If an event occurred for a long period of time, evaluating an event depends on when one's emotion is peak and end moment.

In an aspect of mood congruency, information which is in the same valence with one's present mood would be more perceived, memorized, retrieved and utilized to judge and make a decision than mood incongruent information.

### Chronicle of a Death Foretold

The novel written by Garcia Márquez (1981), Chronicle of a death foretold, came from his experience of seeing a murder case. In a small village near coast, Vicario brother decided to kill Nasar with unproven reason that Nasar raped their sister Angela and then regain their family's honor. The brother mentioned what the time, location, and the reason why they will kill him. However, no one told the foretold danger to Narsar. As a result, the village became full of honor and revenge, violence and indifference, false witness and misunderstanding, as the story drove into tragedy.

For the characteristics of this novel (e.g. honor killing, indifference of the crowd, narrative style, and non-sequential story telling), it has been studied for a long time.

### Sentimental Analysis

Sentimental analysis is a field of text mining, which consists of extracting specific emotional expressions from the text. While text mining is aimed towards finding a designated topic where the text is included, sentimental analysis model tries to find emotional state of the person who wrote the document (Pang & Lee, 2008). Analyzing product reviews can be an example of such approach. They include positive or negative words about the product. Sentimental analysis is the process of automatic classification or scoring of those data, and then we can apply the result to make the product more attractive.

The goal of sentimental analysis is to determine the polarity and the intensiveness of data. Its algorithms focus on word used to show their favors rather than the structure of the words themselves. At first, the algorithm divides the document by polarity, and learns the meaning of words which includes each polarity. The next step is the calculation of scores using an algorithm, applying a positive for words with positive meaning, and vice versa.

In the field of Natural Language Process (NLP), the sentimental analysis problem is a hard to achieve both high accuracy and performance. One of the most commonly used dataset in sentimental analysis is Stanford Sentient Treebank which consists of movie reviews labeled 5 classes (very negative, negative, neutral, positive, very positive) (Socher et al., 2013). The recent model handling the dataset is Dynamic Memory Network (Kumar et al., 2015). Although the model performs well in binary classification (positive or negative), its fine-grained accuracy is only 51.2%, which is considerably low.

### Convolutional Neural Network

A Convolutional Neural Network (LeCun & Bengio, 1995) is a hierarchical model, developed from an idea of human's neural network, and is used in various pattern recognition problems. They operate with two layers, convolutional layer and subsampling layer (also known as max-pooling layer), in turn. Finally, it classifies data through fully-connected layer. Convolutional layer applies filter banks to input images through 2D-filtering. Subsampling layer extracts a local maximum value from input images, which is then mapped to 2D-images. Widening the region gradually, the layer repeats down-sampling. Lastly, using fully-connected layer, a model learns through back-propagation throughout the process to minimize input-out errors.

CNN has been widely used to solve visual problems (e.g. face recognition, hand-writing recognition), but recently it is also being implemented in NLP.

CNN on NLP, the first layer makes word vectors using a lookup table. Considering each word as a pixel, CNN represents each document by (|document|) × 1 vectors where the size of channel is equal to the number of extracted words per document. Then, the model progresses identically to a case of images, which uses the word vectors as feature vectors.

## Transfer Learning

Transfer learning is a method to convey knowledge from similar tasks, which are related to the target task (Pan & Yang, 2010). As with the process of human learning, it has advantages the training requires fewer amounts of data because of employing the knowledge from similar problems.

Transfer learning generally takes three steps: First, the framework learns from the source. Second, the framework transfers the knowledge from source to target, and the last is to learn from the target.

While traditional machine learning concentrates solely on solving a problem using training data in the same domain, transfer learning is not constrained in a specific domain.

# Data

## Experimental Data

### Participants Information

The general information of participants is as follow:

Table 1: The distribution of age of participants

| Age | N | % |
|---|---|---|
| 20 | 19 | 29.69 |
| 21 | 13 | 20.31 |
| 22 | 6 | 9.38 |
| 23 | 8 | 12.5 |
| 24 | 5 | 7.81 |
| 25 | 5 | 7.81 |
| 26 | 5 | 7.81 |
| 27 | 2 | 3.13 |
| 30 | 1 | 1.56 |

All of participants were university students in Republic of Korea. Evenly distributed participants were sampled; 39 (60.94%) male and 25 (39.06%) female and 31 (48.44%) students majored Liberal Arts while the others (51.56%) majored in Science.

### Personal Factors

The psychological status; LOT, SS, CSES, PANAS, CES-D, of the undergraduate students is presented in table 2.

Table 2: Psychological Status of participants

| Characteristic | N | % | Mean (SD) | Range |
|---|---|---|---|---|
| LOT | | | 22.313 (2.241) | 14-25 |
| Optimistic (>17) | 50 | 78.13 | | |
| Pessimistic (<17) | 8 | 12.5 | | |
| Neutral (=17) | 6 | 9.38 | | |
| SS | | | 51.016 (9.505) | 26-93 |
| Family | | | 17.619 (6.439) | 4-60 |
| Friend | | | 16.762 (3.609) | 6-20 |
| Significance | | | 16.635 (3.076) | 8-20 |
| CSES | | | 42.141 (7.586) | 22-60 |
| PANAS | | | | |
| Positive | | | 33.594 (6.592) | 22-50 |
| Negative | | | 28.828 (7.629) | 15-46 |
| CES-D | | | 26.635 (12.207) | 5-50 |
| Depression (>=21) | 41 | 65.08 | | |
| Non-depression (<21) | 22 | 34.92 | | |

## Train Data

We collected movie review data from Naver, the most popular Korean portal site. We have implemented web crawling tool using Scrappy, a Python package. The site constrains to scrap that could crawl only 1,000 pages in a day so we had to collect reviews frequently.

The review data includes the information of titles, scores, comments on the movie. The distribution of scores in the review data is presented in table 3.

Table 3: The distribution of scores of the scrapped review data

| Score | N | % |
|---|---|---|
| 1 | 4,610 | 12.05 |
| 2 | 511 | 1.34 |
| 3 | 452 | 1.18 |
| 4 | 497 | 1.30 |
| 5 | 997 | 2.61 |
| 6 | 1,287 | 3.36 |
| 7 | 2,157 | 5.64 |
| 8 | 3,706 | 9.69 |
| 9 | 4,426 | 11.57 |
| 10 | 19,622 | 51.28 |

## Method

The subjects of the experiment were undergraduate students (N=64) enrolled in "Introduction to Cognitive Science" class in Konkuk University, Republic of Korea, during 2014 fall semester. 64 students are supported in the beginning, the end, only 55 remained.

We examined depression (Center for Epidemiological Studies – Depression, Radloff, 1977), present positive affective and negative affective (Positive Affective and Negative Affective Schedule, Watson et al., 1988), life orient (Life Orient Test, Scheier & Carver, 1985), self-awareness (Core Self Evaluation Scale, Judge et al., 2003) and social factor (Social Support, Sarason et al., 1983)

LOT is question set used to score personal positive or negative characteristics. Its threshold score is 17. If the score is higher than 17, then a person is considered optimistic. On the other hand, it the score is lower than 17, then a person is considered pessimistic. SS checks the recognition of social support and satisfaction (Sarason et al., 1983). It first requires the list all related people, and it also requires rate satisfaction when interacting with those people. CSES consists of 12 questions which measure their have self-efficacy. PANAS measures positive affection (interested, excited, strong, enthusiastic, proud, alert, inspired, determined, attentive, and active) and negative affection (distressed, upset, guilty, scared, hostile, irritable, ashamed, nervous, jittery, and afraid). CES-D score shows how depression is severe during specific periods. The cut-off point for depression, 16 is suggested by Radloff, whereas Cho et al. (1993) suggested 21 for epidemiological studies for Koreans.

Participants first completed their psychological state; CES-D, PANAS as well as personal factors; LOT, CSES, SS. As for their assignments, we gave them a book, 'Chronicle of a Death Foretold' (Gabriel García Márquez, 1981) to read carefully, announcing that they will take a quiz in the following week. In the following week, the students were asked to summarize the content of the book.

To construct our sentimental analysis model, we scrapped movie review data (as stated above) and implemented a simple sentimental analysis model based on convolutional neural network. We first trained the model using the movie data. And then, we input the summaries of the book to our model to evaluate the emotional expressions included in their summaries.

We have obtained the concept from transfer learning, which is that the knowledge from similar tasks can be used for doing target tasks. We hypothesis that movie reviews which have emotional words as well as the intense of the word which is presented as a form of scores, from 1 to 10, can be sources to evaluate emotional contents of the summaries.

Our sentimental analysis model based on convolutional neural network is presented in Figure 2.

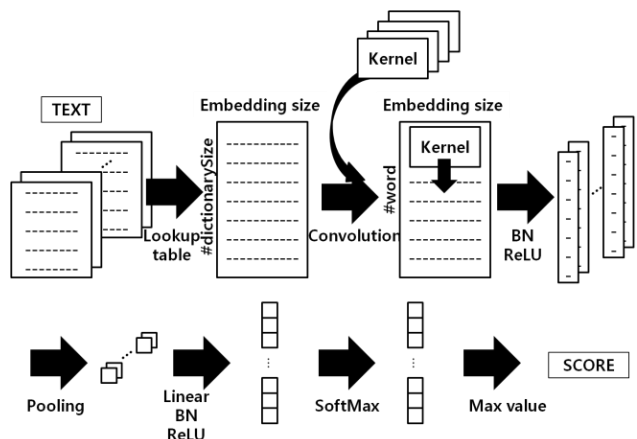

Figure 2: Convolutional neural network Architecture

Unlike other conventional sentimental analysis model, we considered it as a classification problem. We first constructed a word dictionary to count top-n frequency of the words in the data and represented it in vector format. After that, we have slidden a kernel over source data, in order to learn the parameters in kernels. Then we have performed max-pooling, which extracts the maximum value in the range of pooling size the output. Passing linear module and ReLU function (Nair & Hinton, 2010) with batch normalization (Ioffe & Szegedy, 2015), we choose a score which has the highest probability.

## Results

### Model Accuracy

For an experiment, we collected 38,265 movie review data, allocating the ratio of training set, validation set, and test set is 0.7 : 0.15 : 0.15.

### 10-Level polarity

We divide the score of movie reviews from 1 to 10. The accuracy is presented in Table 4.

Table 4: The accuracy of our model as 10-level polarity

| Test Accuracy (Top-1,3,5) | | |
|---|---|---|
| 0.5394 | 0.6933 | 0.7810 |

### 5-Level polarity

We divide the score of movie reviews from 1 to 5. We merge 1 with 2, 3 with 4, 5 with 6, 7 with 8, and 9 with 10 as a group. The accuracy is presented in Table 5.

Table 5: The accuracy of our model as 5-level polarity

| Test Accuracy (Top-1,3,5) | | |
|---|---|---|
| 0.6584 | 0.7992 | - |

**3-Level polarity**

We divide the score of movie reviews as positive (1 to 3), neural (4 to 7), and negative (8 to 10). The accuracy is presented in Table 6.

Table 6: The accuracy of our model as 3-level polarity

| Test Accuracy (Top-1,3,5) | | |
|---|---|---|
| 0.7598 | - | - |

**Score information**

We take 4 different scoring approaches: (i) scoring from 1 to 10 as a paragraph. (ii) scoring from 1 to 5 as a paragraph. (iii) scoring from 1 to 10 as sentences (iv) scoring from 1 to 5 as sentences.

In first approach, 39 (70.91%) summaries were 1 point, 2 of them (3.64%) 3 points, 1 of them (1.82%) 5 points, 1 of them (1.82%) 6 points, 4 of them (7.27%) 8 points, and 8 of them (14.55%) 10 points.

In second, 2 (3.64%) summaries were 1 point, 7 of them (12.73%) 4 points, and 46 of them (83.64%) 5 points.

The third, 430 (70.03%) sentences of summaries were 1 point, 3 of them (0.49%) 2 points, 110 of them (17.92%) 3 points, 33 of them (5.37%) 5 points, 14 of them (2.28%) 6 points, 10 of them (1.63%) 8 points, 1 of them (0.16%) 9 points, and 13 of them (2.12%) 10 points.

The last, 20 (3.26%) sentences of summaries were 1 point, 34 of them (5.54%) 2 points, 8 of them (1.30%) 3 points, 59 of them (9.61%) 4 points, and 493 of them (80.29%) 5 points.

**Relationship with Psychological Factors**

Under a hypothesis that the scoring is reasonable, we have investigated for a relationship between the scores and psychological factors.

First, we divided the participants by CES-D using cut-off point, 21. For illustration, Para1to10 means the score in the case of (i) described above. Likewise, Para1to5 means the score in the second case, Sen1to10 means the score of (iii), and Sen1to5 means the score of (iv).

Table 7: Scores information when divided by CES-D

| Characteristic | | N* | Mean (SD). | $t$ or $F$-value |
|---|---|---|---|---|
| Para1to10 | Depressed | 34 | 3.676 (3.890) | $F=15.118$ |
| | Non-depressed | 21 | 2.048 (2.519) | $p=.065$ |
| Para1to5 | Depressed | 34 | 4.735 (0.751) | $T=.093$, |
| | Non-depressed | 21 | 4.714 (0.902) | $p=.926$ |
| Sen1to10 | Depressed | 34 | 2.008 (0.823) | $T=-.491$, |
| | Non-depressed | 21 | 2.121 (0.854) | $p=.625$ |
| Sen1to5 | Depressed | 34 | 4.590 (0.399) | $T=.897$, |
| | Non-depressed | 21 | 4.486 (0.446) | $p=.374$ |

We could observe a difference in scores between depressed (M=3.676, SD=3.890) and non-depressed (M=2.048, SD=2.519) in the first approach; $F(52.835)=15.118$, $p=.065$.

Also, We divide participants by PANAS using a threshold as mean value, 61.890.

Table 8: Scores information when divided by PANAS

| Characteristic | | N* | Mean (SD). | $t$ or $F$-value |
|---|---|---|---|---|
| Para1to10 | Higher PANAS | 26 | 2.962 (3.447) | $T=-.185$ |
| | Lower PANAS | 29 | 3.138 (3.603) | $p=.854$ |
| Para1to5 | Higher PANAS | 26 | 4.846 (0.368) | $T=1.039$, |
| | Lower PANAS | 29 | 4.621 (1.049) | $p=.304$ |
| Sen1to10 | Higher PANAS | 26 | 1.819 (0.746) | $T=-2.021$, |
| | Lower PANAS | 29 | 2.259 (0.857) | $p=.048$ |
| Sen1to5 | Higher PANAS | 26 | 4.649 (0.274) | $F=7.008$, |
| | Lower PANAS | 29 | 4.462 (0.500) | $p=.089$ |

We could observe a significant difference in scores between higher PANAS group (M=1.819, SD=0.746) and lower PANAS group (M=2.259, SD=0.857) in the third approach; $t(53)=-2.021$, $p=.048$. In the same vein, the fourth approach shows a significant difference; $F(44.374)=7.008$, $p=.089$.

## Conclusion

Above all, how we define the meaning of scores is an important crossroad to interpret the results. Specifically, 1 point can be interpreted in two ways: very negative or very impassive. Therefore, both valence and intensity have to be taken into consideration when drawing a conclusion with derived scores. When scores were considered as valence, depressed has shown a tendency of representing a story more negatively, when compared to a non-depressed. This can be explained as mood congruency. On the other hand, when scores were considered as intensity, the depressed seemed less expressive than those who are non-depressed. This can be explained as affect regulation. People who scored relatively high in both positive and negative affection, those who has greater affection, have used more negative words than others. In contrast, people with full of emotion have retrieved their memory more expressively than others.

The CES-D results correspond to our previous research (Jo et al., 2016) which states that depressed people tend to overgeneralize their memory to avoid negative experience through affect regulation.

Interesting results were found in PANAS. The results directly prove the relationship between emotion and memory retrieval. However, other psychological factors, LOT, CSES, SS failed to show any significant difference.

We have used 4 different approaches to score summaries from different classification approaches we have taken: 'as a whole' or 'sentence by sentence'. We have wondered that scoring from 1 to 10 is too detailed to distinguish between different levels of emotion and would result in unsatisfactory accuracy. For instance, when taking two people – one who have scored 1 point and the other who have scored 2 points – into consideration, significant difference between two test subjects cannot be drawn. Therefore, we have taken various approaches to eliminate this error.

Limitation and tentative work plan are as follows: the training data is insufficient for our model to classify sentiments in better performance. Second, the movie review data is not reliable because some companies employ part-time jobs to fabricate their movie's review scores. Transfer of learning might not an appropriate approach in this case - movie review to text summary. Considering our model accuracy, finding other sentimental analysis architecture is needed. As a matter of data imbalance, factor analysis method would be suggested.